\documentclass{article}

\usepackage{PRIMEarxiv}

\usepackage[utf8]{inputenc} 
\usepackage[T1]{fontenc}    
\usepackage{hyperref}       
\usepackage{url}            
\usepackage{booktabs}       
\usepackage{amsfonts}       
\usepackage{nicefrac}       
\usepackage{microtype}      
\usepackage{lipsum}
\usepackage{fancyhdr}       
\usepackage{graphicx}       
\graphicspath{{media/}}     

\usepackage{tabularx}
\usepackage{pifont} 
\usepackage{makecell}
\usepackage{array}
\usepackage{ragged2e}
\usepackage[table,xcdraw]{xcolor}
\usepackage{multirow}
\usepackage{arydshln}   
\usepackage{stfloats} 

\usepackage{wasysym} 
\usepackage{geometry}
\usepackage{enumitem}
\usepackage{xcolor}   
\usepackage{tikz}     

\newcommand{\filledsmiley}{%
  \begin{tikzpicture}[scale=0.12, baseline=-0.3ex]
    \draw[fill=yellow!70!yellow, draw=black, line width=0.1pt] (0,0) circle (1); 
    \fill[black] (-0.35,0.3) circle (0.12); 
    \fill[black] (0.35,0.3) circle (0.12);  
    \draw[black, line width=0.4pt] (-0.5,-0.2) arc (210:330:0.58); 
  \end{tikzpicture}%
}

\newcommand{\redcross}{\textcolor{red}{\boldmath$\times$}}
\newcommand{\greencheck}{\textcolor{green!70!black}{\checkmark}}

\definecolor{myblue}{rgb}{0.21,0.49,0.74}

\hypersetup{
hidelinks,
colorlinks=true,
linkcolor=red,
citecolor=myblue
}

\title{SynthVerse: A Large-Scale Diverse Synthetic Dataset for Point Tracking}

\author{
Weiguang\,Zhao$^{1,6,*}$\;\;\,
Haoran\,Xu$^{2,*}$\;\;\,
Xingyu\,Miao$^{3}$\;\;\,
Qin\,Zhao$^{2}$\;\;\,
Rui\,Zhang$^{6,\dagger}$\;\;\,\\
\textbf{Kaizhu\,Huang}$^{5,\dagger}$\;\;\
\textbf{Ning\,Gao}$^{7,9}$\;\;\
\textbf{Peizhou\,Cao}$^{4,9}$\;\;\
\textbf{Mingze\,Sun}$^{8}$\;\;\
\textbf{Mulin\,Yu}$^{9}$\;\;\ \\
\textbf{Tao\,Lu}$^{9}$\;\;\
\textbf{Linning\,Xu}$^{9,10}$\;\;\
\textbf{Junting\,Dong}$^{9,\dagger,\ddagger }$\;\;\,
\textbf{Jiangmiao\,Pang}$^{9}$
\\
$^{1}$University of Liverpool  \quad $^{2}$Zhejiang University \quad $^{3}$Durham University \quad $^{4}$Beihang University   \\ \quad $^{5}$Duke Kunshan University \quad $^{6}$Xi'an Jiaotong-Liverpool University   \quad $^{7}$Xi'an Jiaotong University\\  \quad $^{8}$Tsinghua University \quad $^{9}$Shanghai AI Laboratory  \quad $^{10}$The Chinese University of Hong Kong\\
Project page: \url{https://weiguangzhao.github.io/SynthVerse/}
}

\begin{document}
\maketitle

\begingroup
\renewcommand{\thefootnote}{\fnsymbol{footnote}}
\footnotetext[3]{Project lead.}
\footnotetext[1]{Equal contribution.}
\footnotetext[2]{Corresponding author.}
\renewcommand\thefootnote{}\footnotetext{This work was completed during an internship at Shanghai AI Lab.}
\endgroup

\begin{abstract}
Point tracking aims to follow visual points through complex motion, occlusion, and viewpoint changes, and has advanced rapidly with modern foundation models. Yet progress toward general point tracking remains constrained by limited high-quality data, as existing datasets often provide insufficient diversity and imperfect trajectory annotations. To this end, we introduce SynthVerse, a large-scale, diverse synthetic dataset specifically designed for point tracking. SynthVerse includes several new domains and object types missing from existing synthetic datasets, such as animated-film-style content, embodied manipulation, scene navigation, and articulated objects.  SynthVerse substantially expands dataset diversity by covering a broader range of object categories and providing high-quality dynamic motions and interactions, enabling more robust training and evaluation for general point tracking.  In addition, we establish a highly diverse point tracking benchmark to systematically evaluate state-of-the-art methods under broader domain shifts. Extensive experiments and analyses demonstrate that training with SynthVerse yields consistent improvements in generalization and reveal limitations of existing trackers under diverse settings.
\end{abstract}


\begin{figure}[ht]
\centering
  \includegraphics[width=\textwidth]{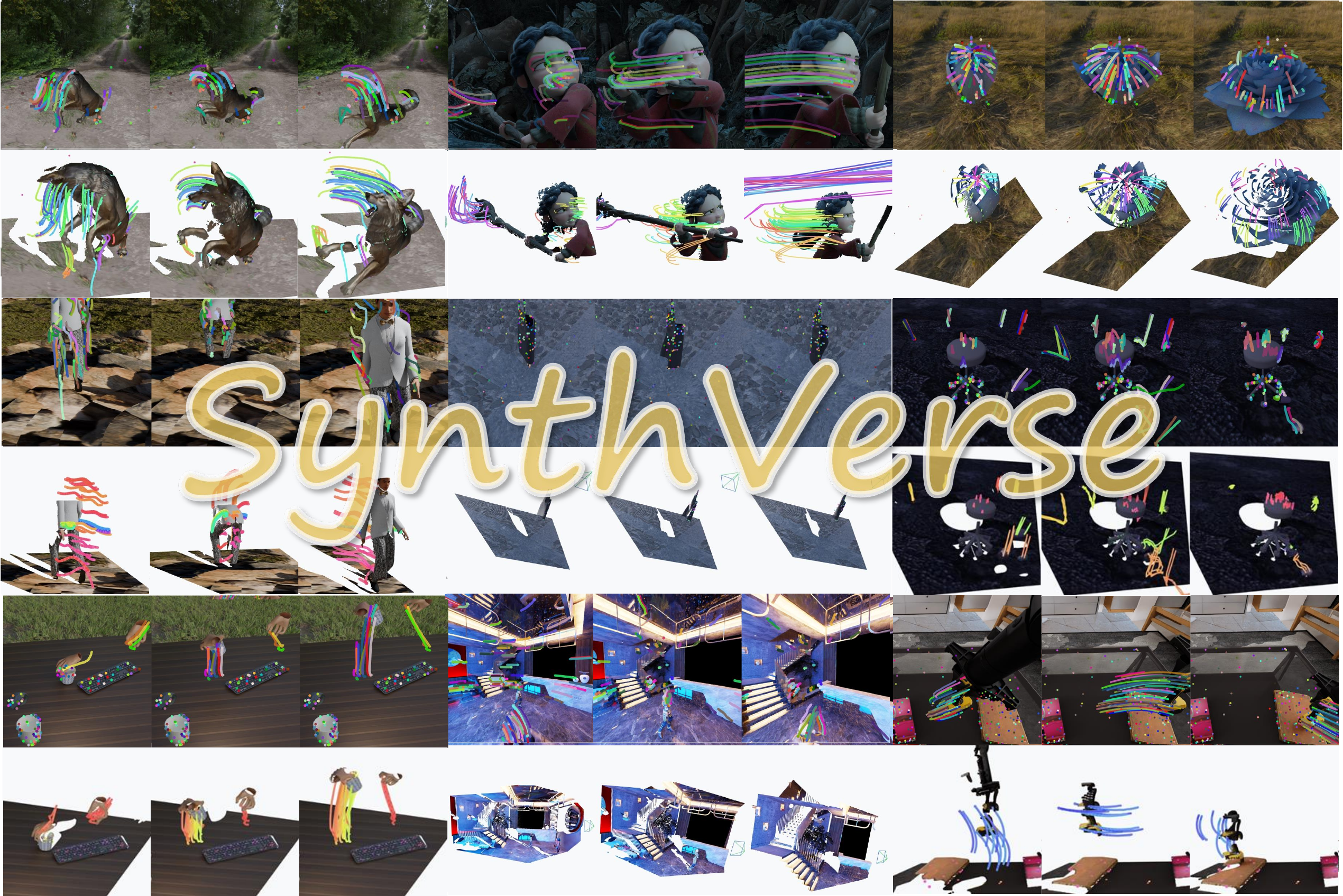}
  \caption{Our SythnVerse Dataset}
  \label{fig:teaser}
\end{figure}

\section{Introduction}
Point tracking~\cite{cotracker3,ngodelta,ngo2025deltav2,wang2025scenetracker} is a fundamental primitive for scene understanding, 4D reconstruction, and robot perception, serving as a key building block for reasoning about motion and correspondence over time. Despite rapid progress driven by modern foundation models, achieving robust general tracking under diverse motions and scenes remains highly challenging, especially under distribution shifts across domains and object types. A key bottleneck is the lack of large-scale, high-quality, and sufficiently diverse training data~\cite{tian2025interndata,zhou2025omniworld,wang2025pi,li2025sekai} with reliable trajectory annotations.

Building a large-scale real-world point tracking dataset is notoriously difficult, since accurate tracking ground truth is inherently hard to obtain in real scenes~\cite{koppula2024tapvid}. This challenge becomes even more severe for 3D point tracking, where supervision requires not only consistent 2D correspondences but also accurate depth, camera poses, and temporally stable 3D trajectories under occlusions, motion blur, and sensor noise. Real-world annotations often rely on costly human labeling or optimization-based post-processing, both of which are time-consuming and inevitably introduce errors. In contrast, synthetic data provides a scalable alternative. The assets and scenes are fully controllable, enabling efficient large-scale rendering, while precise ground-truth trajectories can be directly derived from the underlying physical states.

Consequently, recent point tracking research has increasingly adopted synthetic datasets as the primary source for model training. Existing datasets and benchmarks~\cite{zheng2023pointodyssey,greff2022kubric, karaev2023dynamicstereo,koppula2024tapvid} still offer limited coverage for general point tracking. As summarized in Tables~\ref{tab:ptdb} and~\ref{tab:coe}, existing synthetic datasets exhibit clear limitations in scale and breadth, with restricted diversity in scene composition, motion patterns, and object categories. Furthermore, most datasets are constructed within a single simulation/rendering pipeline and predominantly adopt allocentric camera setups, leaving embodied interactions, navigation-style dynamics, and egocentric observations insufficiently represented. These constraints lead to biased training distributions and exacerbate distribution shifts across domains, underscoring the need for a substantially larger and more diverse synthetic dataset and benchmark for general point tracking.

To this end, we introduce SynthVerse, a large-scale and diverse synthetic dataset specifically designed for general point tracking. SynthVerse, as depicted in Fig.~\ref{fig:teaser}, substantially expands both the scale and content diversity of existing synthetic corpora, comprising 5,816K training frames and 48K sequences, while additionally covering object types that have not been systematically supported before, including articulated and deformable objects, along with rich human and animal motions. In contrast to prior datasets built on a single pipeline and dominated by allocentric captures, SynthVerse is generated via a cross-platform pipeline integrating Blender and Isaac Sim, supports both egocentric and allocentric viewpoints, and spans several previously missing yet critical domains—such as animated-film-style content, embodied manipulation, navigation, and hand interactions—thereby enabling more realistic dynamic patterns for training.

We further establish a highly diverse point tracking benchmark based on SynthVerse, enabling systematic evaluation of state-of-the-art trackers under a broad range of domains, object types, and motion patterns. This benchmark reveals the limitations of existing 3D point tracking models when confronted with domain shifts, and highlights how their performance is influenced by the training data distribution. Extensive experiments further demonstrate the effectiveness of SynthVerse as a scalable training source, consistently improving model robustness and generalization across diverse settings. In summary, our contributions are as follows:

\begin{itemize}[leftmargin=8pt]

\item We introduce SynthVerse, a large-scale and diverse synthetic dataset for general point tracking, together with a cross-platform data generation pipeline to produce reliable trajectory annotations with broad domain and object coverage.

\item We build diverse point tracking benchmarks to evaluate trackers under broader distribution shifts and challenging dynamics.

\item We conduct extensive evaluations and systematic analyses of state-of-the-art point tracking methods, demonstrating the effectiveness of SynthVerse for training and benchmarking general point tracking.

\end{itemize}

\section{Related Work}

\subsection{Point Tracking Dataset and Benchmark}

Existing mainstream point tracking datasets are summarized in Table~\ref{tab:ptdb}. Considering it is difficult to obtain ground-truth point trajectories from real-world captured scenes, most datasets~\cite{joo2015panoptic,pan2023aria,koppula2024tapvid} in this category only provide test sets, whose annotations are produced manually. DriveTrack~\cite{balasingam2024drivetrack} estimates point trajectories indirectly by fitting trajectory transformations from bounding boxes. This strategy is only applicable to rigid objects and can introduce noticeable trajectory errors. In contrast, synthetic data enables access to complete object states in simulation, making it possible to derive accurate point trajectories. 

Although several high-quality synthetic datasets~\cite{greff2022kubric,zheng2023pointodyssey,karaev2023dynamicstereo} have been proposed, their simulated scenes tend to be relatively limited in diversity. We provide a detailed comparison between our dataset and existing synthetic datasets in Table~\ref{tab:coe}. Specifically, Kubric~\cite{greff2022kubric} is the pioneering work to systematically leverage large-scale synthetic data for point tracking. It primarily focuses on rigid-body free-fall scenarios and does not provide point tracking trajectories for articulated objects or deformable objects. PointOdyssey~\cite{zheng2023pointodyssey} provides a long-term point tracking dataset and expands the content coverage by including animals and humans, along with a subset of deformable-object scenarios. LSFOdyssey~\cite{wang2025scenetracker} applies data augmentation to the original PointOdyssey data (e.g., random flipping, spatial transformations, and color transformations), and is therefore not listed separately in Table~\ref{tab:coe}. Dynamic Replica~\cite{karaev2023dynamicstereo}  further incorporates more human and animal content; however, the number of such sequences remains relatively small. Overall, existing synthetic point-tracking datasets remain limited in scale and exhibit insufficient diversity in both object categories and scene types.

In contrast, our dataset is substantially larger (5816K training frames, 59K test frames, and 48K sequences) and covers a broader spectrum of content (approximately 1K articulated-object sequences, 6K deformable-object sequences, 75 animal sequences, and 4K human sequences). Moreover, by combining Blender and Isaac Sim, our pipeline supports both egocentric and allocentric views and spans diverse scene types, including embodied manipulation, indoor navigation, hand interaction, and animated-film-style scenarios, providing a more comprehensive benchmark for point tracking.

\begin{table}[h]
\centering
\caption{Point Tracking Dataset and Benchmark. }
\label{tab:ptdb}
\setlength{\tabcolsep}{1.5pt}
\renewcommand{\arraystretch}{1.15}

\resizebox{0.70\linewidth}{!}{%
\begin{tabular}{lcclcl}
\toprule
Dataset & Train Set & Test Set & Type & Total Frames & Type \\
\midrule
Pstudio~\cite{joo2015panoptic}        & \redcross  & \greencheck  & Real           & 15K   & Human \\
ADT~\cite{pan2023aria}             & \redcross  & \greencheck  & Real           & 571K  & Indoor \\
DriveTrack~\cite{balasingam2024drivetrack}      & \greencheck  & \greencheck  & Real           & 756k  & Drive \\
TAPVid-3D~\cite{koppula2024tapvid}       & \redcross  & \greencheck  & Real/Synthetic & 600k  & Diverse \\
Kubric~\cite{greff2022kubric}           & \greencheck  & \greencheck  & Synthetic      & 62K   & Object-Falling \\
PointOdyssey~\cite{zheng2023pointodyssey}     & \greencheck  & \greencheck  & Synthetic      & 192K  & Human Animal \\
Dynamic Replica~\cite{karaev2023dynamicstereo} & \greencheck & \greencheck  & Synthetic      & 163K  & Human Animal \\
\rowcolor{gray!20} SynthVerse (Ours)           & \greencheck & \greencheck  & Synthetic      &   5816K    & Diverse \\
\bottomrule
\end{tabular}%
}
\end{table}

\begin{table}[h]
\centering
\caption{Comparisons between ours and existing synthetic datasets}
\label{tab:coe}
\footnotesize
\setlength{\tabcolsep}{2.5 pt}
\renewcommand{\arraystretch}{1.12}
\resizebox{\textwidth}{!}{%
\begin{tabularx}{\textwidth}{lccccccccc}
\toprule
{Dataset} &
\makecell{{Training}\\{Frames}} &
\makecell{{Test}\\{Frames}} &
\makecell{{Sequence}\\{Count}} &
\makecell{{Articulated}\\{Object}} &
\makecell{{Deformable}\\{Object}} &
{Animal} &
{Human} &
{Pipeline} &
{View} \\
\midrule

Kubric~\cite{greff2022kubric}          & 60K  & 2.4K & 2.5K & 0   & 0   & 0  & 0   & Blender & Allo-centric  \\

PointOdyssey~\cite{zheng2023pointodyssey}    & 166K & 26K  & 80   & 0   & 49  & 7  & 42  & Blender & Allo-centric \\

Dynamic Replica~\cite{karaev2023dynamicstereo} & 145K & 18K  & 484  & 0   & 388 & 13 & 375 & Blender & Allo-centric  \\
\midrule
\textbf{SynthVerse (Ours)} &  \textbf{5816K}  & \textbf{59K} & \textbf{48K} & 
\textbf{1K} & \textbf{6K} & \textbf{75} & \textbf{4K} &
\makecell[c]{\textbf{Blender} \\ \textbf{Isaac Sim}} &
\makecell[c]{\textbf{Ego-centric} \\ \textbf{Allo-centric}} \\
\bottomrule
\end{tabularx}
}
\end{table}

\subsection{Point Tracking Methods}
2D point tracking aims to follow given pixel points over time in a video and predict their corresponding 2D locations at each frame, together with visibility or occlusion. As the pioneer work, PIP~\cite{harley2022particle} formulates pixel tracking as a long-range trajectory estimation problem, combining dense cost maps, iterative optimization, and learned temporal priors to track points consistently even under occlusion. TAPIR~\cite{doersch2023tapir} combines per-frame matching with temporal refinement to achieve state-of-the-art accuracy and real-time efficiency for tracking any point in long, high-resolution videos. MVTracker~\cite{rajivc2025multi} achieves robust and efficient tracking of arbitrary points in dynamic scenes by fusing multi-view features into a unified 3D point cloud and leveraging kNN-based correlation with a transformer architecture. The CoTracker~\cite{karaev2024cotracker,cotracker3} line of work further advances the state of the art in 2D point tracking.

Moreover, SpatialTracker~\cite{xiao2024spatialtracker} lifts 2D pixels into 3D space for tracking and leveraging a triplane representation with as-rigid-as-possible constraints to effectively handle occlusions and complex motions. Furthermore, SpatialTrackerV2~\cite{xiao2025spatialtrackerv2} provides a unified end-to-end differentiable framework that achieves efficient and high-accuracy 3D point tracking from monocular videos by jointly modeling scene geometry, camera ego-motion, and object motion. DELTA enables efficient 3D tracking of every pixel in any video, running over 8x faster than existing methods while achieving state-of-the-art accuracy in both 2D and 3D dense tracking. In addition, TAPIP3D~\cite{zhang2025tapip3d} achieves robust long-term 3D point tracking by representing videos as camera-stabilized 3D feature clouds and introducing a 3D Neighborhood-to-Neighborhood (N2N) attention mechanism to track points in world-centric coordinates. D4RT~\cite{zhang2025efficiently} introduces a flexible query-based decoding mechanism that enables efficient, independent 3D position probing for any point in space and time, unifying dynamic 4D reconstruction and tracking into a single, scalable feedforward interface.

\begin{figure*}[ht]
  \includegraphics[width=\textwidth]{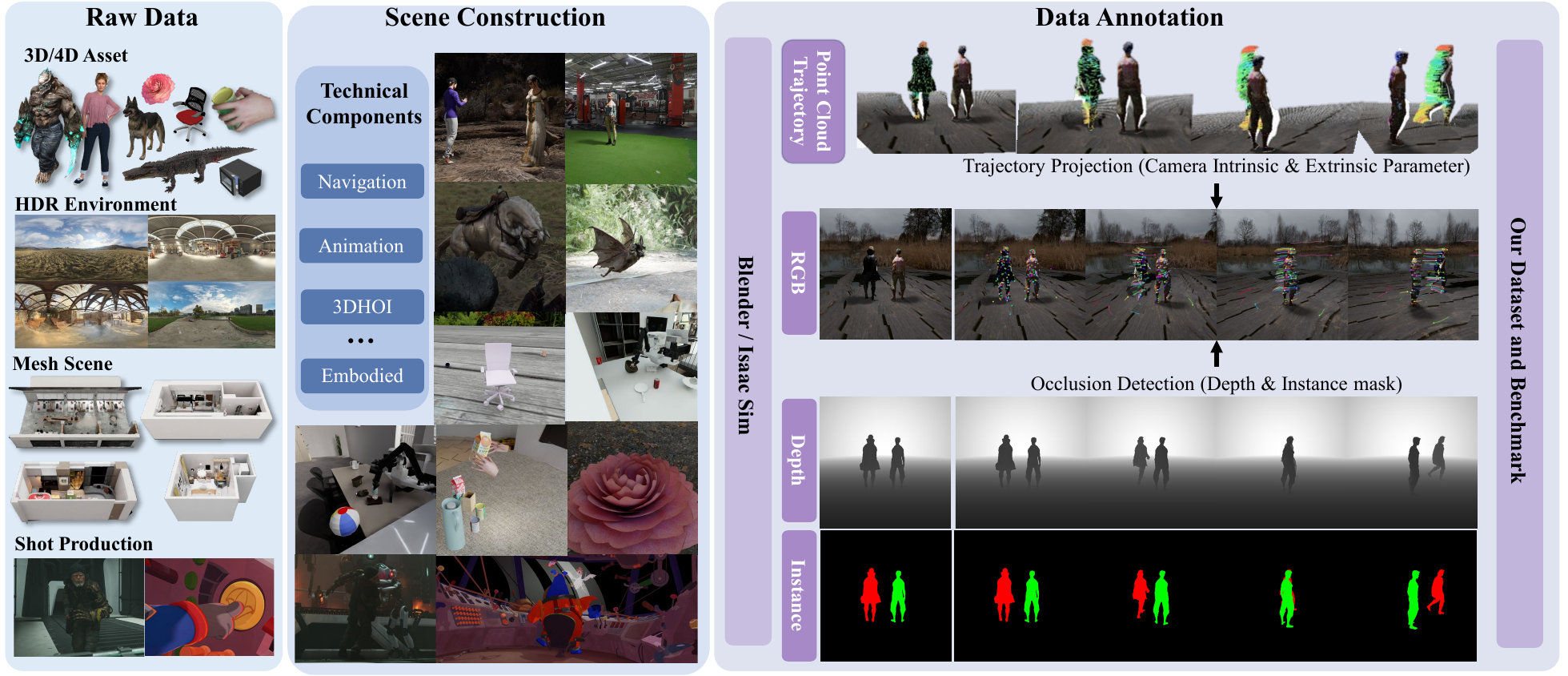}
  \caption{Data Generation Pipeline. Shot Production denotes publicly released shot-level production project files from selected clips of animated films. Technical Components refer to a collection of technical methods and tool modules used during scene construction to facilitate scene layout, motion setup, and related production controls.}
  \label{fig:pipeline}
\end{figure*}

\section{Dataset Pipeline}

\subsection{Overview of the Data Generation Pipeline}
As illustrated in Fig.~\ref{fig:pipeline}, we develop a unified data generation pipeline for SynthVerse, which starts from raw data acquisition and scene construction, and then produces consistent annotations. The pipeline is implemented across Blender and Isaac Sim, enabling scalable synthesis and reliable supervision. 

We curate diverse raw resources to support broad visual and physical variations, including 3D/4D assets, HDR environment maps, mesh-based scenes, and shot-level production content. These sources jointly provide rich geometry, appearance, illumination, and motion priors, serving as the foundation for composing diverse technical components such as embodied scenarios, navigation, animation, and 3D human–object interaction (3DHOI) in a unified pipeline.

Scene Construction. Based on the curated resources, we construct dynamic scenes within the Blender/Isaac Sim workflow and organize the generation process into multiple technical components, including Embodied~\cite{gao2025genmanip}, Navigation~\cite{zhong2025internscenes}, Animation, and 3D Human–Object Interaction (3DHOI)~\cite{banerjee2025hot3d}. During scene construction, we incorporate several specialized techniques—for example, VLA-driven embodied behaviors are used to generate interaction-rich motions, and 3DHOI scenes are reconstructed in Blender to reproduce realistic human–object contacts and dynamics. Specifically, we further organize the generated content into scene-level data with full backgrounds and instance-level assets without backgrounds. For scene-level data, we preserve the original scene layout and camera configuration whenever possible to maintain realistic composition and motion patterns. For example, we reuse Isaac Sim scenes for embodied manipulation and augment them with multi-view cameras, adopt production camera setups for animated-film shots, and generate camera trajectories for navigation sequences while introducing dynamic actors to enrich motion and occlusion diversity.

During rendering, we record multi-modal information, including RGB images, depth maps, and instance masks, and generate accurate 3D point trajectories from the underlying scene states. Specifically, we first obtain point trajectories in 3D space, and then project them to the image plane using camera intrinsics and extrinsics to derive the corresponding 2D trajectories. We further determine point visibility via occlusion detection based on depth and instance masks, producing temporally consistent annotations for both training data and evaluation benchmarks.

\begin{table}
    \centering
    \small
    \setlength{\tabcolsep}{2pt}
    \caption{Source Data. Ins., Cam., Traj., and Vis. separately denote instance masks, camera poses, point trajectories, and point visibility. Moreover, \greencheck~indicates that the original dataset already provides the corresponding signal. \redcross~is that the original dataset does not include this information, and we also do not provide it. \protect\filledsmiley~represent that the original dataset lacks this information, but we additionally annotate and incorporate it in our SynthVerse.}
    \resizebox{0.6075\linewidth}{!}{
    \begin{tabular}{clcccccc}
    \toprule
    \textbf{SynthVerse} & \textbf{Raw Data} & \textbf{RGB} & \textbf{Depth} & \textbf{Ins.} & \textbf{Cam.} & \textbf{Traj.} & \textbf{Vis.} \\
    \midrule

    \multirow{1}{*}{\centering Embodied}
      & GenManip~\cite{gao2025genmanip} & \greencheck & \greencheck & \greencheck & \greencheck & \filledsmiley & \filledsmiley \\
    \midrule

    \multirow{1}{*}{\centering Human}
      & TCHCDR~\cite{TCHCDR} & \greencheck & \greencheck & \filledsmiley & \greencheck & \filledsmiley & \filledsmiley \\
    \midrule

    \multirow{2}{*}{\centering Animal}
      & Truebones~\cite{TruebonesZoo} & \greencheck & \filledsmiley & \filledsmiley & \filledsmiley & \filledsmiley & \filledsmiley \\
      & AnyTop~\cite{gat2025anytop} & \greencheck & \filledsmiley & \filledsmiley & \filledsmiley & \filledsmiley & \filledsmiley \\
    \midrule

    \multirow{5}{*}{\centering Objects}
      & OmniObject3D~\cite{wu2023omniobject3d} & \greencheck & \filledsmiley & \filledsmiley & \filledsmiley & \filledsmiley & \filledsmiley \\
      & PartNet-M~\cite{Xiang_2020_SAPIEN} & \greencheck & \filledsmiley & \filledsmiley & \filledsmiley & \filledsmiley & \filledsmiley \\
      & Infinite-M~\cite{lian2025infinite} & \greencheck & \filledsmiley & \filledsmiley & \filledsmiley & \filledsmiley & \filledsmiley \\
      & BlendSwap~\cite{blenderswap} & \greencheck & \filledsmiley & \filledsmiley & \greencheck & \filledsmiley & \filledsmiley \\
      & Blender-Demo~\cite{blenderdemo} & \greencheck & \filledsmiley & \filledsmiley & \greencheck & \filledsmiley & \filledsmiley \\
    \midrule

    \multirow{1}{*}{\centering Film}
      & Blender-Studio~\cite{blenderstudio} & \greencheck & \filledsmiley & \redcross & \greencheck & \filledsmiley & \filledsmiley \\
    \midrule

    \multirow{2}{*}{\centering Navigation}
      & InternScenes~\cite{zhong2025internscenes} & \greencheck & \greencheck & \redcross & \greencheck & \filledsmiley & \filledsmiley \\
      & Mixamo~\cite{mixamo} & \greencheck & \filledsmiley & \redcross & \filledsmiley & \filledsmiley & \filledsmiley \\
    \midrule

    \multirow{2}{*}{\centering Interaction}
      & Hot3D~\cite{banerjee2025hot3d} & \greencheck & \filledsmiley & \greencheck & \greencheck & \greencheck & \filledsmiley \\
      & HTML~\cite{qian2020html} & \greencheck & \filledsmiley & \filledsmiley & \filledsmiley & \filledsmiley & \filledsmiley \\
    \bottomrule
    \end{tabular}
    }
    \label{tab:data}
\end{table}

\subsection{Data Acquisition}
In this section, we describe how we acquire and curate the raw data used to build SynthVerse. As summarized in Table~\ref{tab:data}, we collect data from multiple sources and organize them into several asset categories to support diverse visual appearances and physical dynamics. This acquisition strategy provides a scalable foundation for subsequent scene instantiation and annotation, enabling broad coverage across domains and object types for general point tracking.

\noindent \textbf{Embodied} Our embodied manipulation data is primarily collected from GenManip~\cite{gao2025genmanip}, the  VLA-driven embodied manipulation systems that generate high-quality interaction trajectories and dynamic motions across diverse scenarios. We control the robot to perform diverse actions using text-based manipulation instructions, and capture each scene with four cameras: an orbiting camera, a top-down camera, and two hand-mounted cameras. All scene setups are designed to closely resemble real-world environments, ensuring the realism and overall quality of the collected data.

\noindent \textbf{Human} Compared to prior datasets, we curate high-quality human dynamic assets from multiple sources. In particular,  TCHCDR~\cite{TCHCDR} provide detailed, high-fidelity human character models. We collect approximately 20K human characters and 2K motion sequences, and combine them to construct highly diverse human point tracking data. In addition, we incorporate a wide range of humanoid assets from Mixamo~\cite{mixamo}, including non-human humanoids and stylized characters, to further enrich body-shape motion distributions.

\noindent \textbf{Animal} SynthVerse includes point tracking data for 75 animal species (e.g., cats, deer, gazelles, and foxes). Each species is associated with 20+ motion patterns, such as attacking, forward jumping, and swaying, providing broad coverage of animal dynamics. The assets  are sourced from Truebones ZOO~\cite{TruebonesZoo} and AnyTop~\cite{gat2025anytop}.

\noindent \textbf{Object} We collect assets from three categories: rigid bodies, articulated objects, and deformable objects. The rigid-body assets are sourced from the large-scale 3D object dataset OmniObject3D~\cite{wu2023omniobject3d}. Articulated objects constitute a new component of our dataset that has not been systematically covered in prior synthetic point tracking datasets. We collect URDF assets with multi-joint (hinged) structures from PartNet-Mobility~\cite{Xiang_2020_SAPIEN} and Infinite Mobility~\cite{lian2025infinite}, convert them into USD, and import them into Isaac Sim for physics-based simulation and rendering. The aforementioned Human and Animal categories primarily consist of deformable objects. In addition, we further collect deformable assets such as flowers and garments from Blender Demo~\cite{blenderdemo} and BlendSwap~\cite{blenderswap} to enrich scene and appearance diversity.

\noindent \textbf{Animated film} We additionally leverage shot-level production projects released by Blender Studio~\cite{blenderstudio}, which are publicly available for non-commercial use. Animated-film scenes contain a wide spectrum of assets and complex dynamics—ranging from diverse objects and characters to challenging phenomena such as fluids—offering rich motion patterns and occlusion configurations. We integrate these projects into our pipeline and generate point tracking annotations for the relevant visible elements.

\noindent \textbf{Navigation} SynthVerse also contains indoor navigation data, where a camera traverses indoor rooms and moves through the environment. The indoor scene distribution and assets are sourced from InternScenes~\cite{zhong2025internscenes}, and the navigation trajectories are generated using NavDP~\cite{cai2025navdp}. Since InternScenes largely consists of static indoor environments, we additionally introduce dynamic human actors along the navigation trajectories to increase motion complexity and occlusion variation. The human assets and motions are sourced from Mixamo~\cite{mixamo}. we convert the Mixamo FBX assets into USD and import them into Isaac Sim for simulation and rendering.

\noindent \textbf{Hand interaction} We leverage the object state trajectories, 3D assets, and hand state trajectories provided by Hot3D~\cite{banerjee2025hot3d} to reconstruct interactive scenes in Blender and re-render them to generate hand-object interaction data. This component introduces a new data construction paradigm and provides high-quality point tracking annotations for hand-centric interactions.

\noindent \textbf{Others}  To enhance visual diversity and enable domain randomization in SynthVerse, we curate 510 HDR environment maps for lighting and background illumination in both Blender and Isaac Sim. We further collect 4,032 texture images to vary surface appearances (e.g., tabletops and fabrics) via texture replacement, enriching the overall appearance distribution.

\begin{figure}[ht]
\centering
\includegraphics[width=0.50\textwidth]{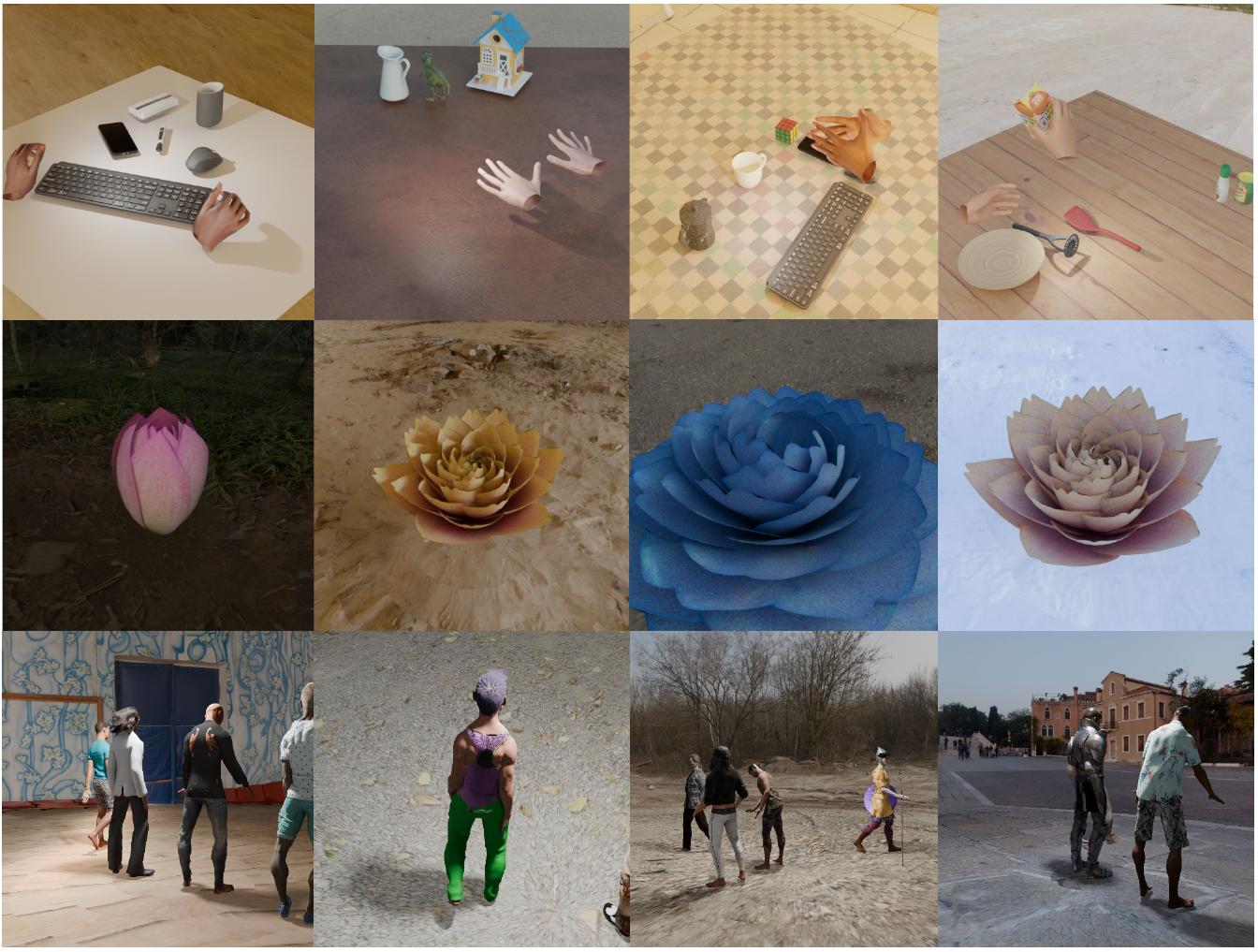} 
\caption{Data Augmentation. The first row shows randomized texture replacements for hands and tabletops. The second row illustrates randomization of flower-petal materials (e.g., base color/albedo). The third row demonstrates camera and environment augmentation, including randomizing camera FOV and HDR environment maps.}
\label{fig:da}
\end{figure}

\subsection{Scene Construction} 
Our collected assets include both scene-level content with full backgrounds (e.g., embodied manipulation, animated-film shots, and indoor navigation) and instance-level assets without backgrounds (e.g., humans, animals, hand interactions, and articulated objects). For scene-level assets, we preserve the original scene layout and camera configuration whenever possible. Specifically, for embodied manipulation, we directly use the original Isaac Sim scenes and add multiple cameras for rendering and data capture. For animated-film content, we adopt the original production camera setups and render the shots as-is to collect trajectories. For indoor navigation, we generate camera traversal paths using NavDP~\cite{cai2025navdp} and introduce dynamic characters along the trajectories, allowing the camera to move above the actors to induce richer motion and occlusion patterns.

For instance-level assets without backgrounds, following the spirit of Kubric~\cite{greff2022kubric} and PointOdyssey~\cite{zheng2023pointodyssey}, we create an HDR-based surround environment in Blender and Isaac Sim and render with multi-view camera setups (four fixed-view cameras plus one orbiting camera). We further standardize motion scheduling for different asset types: for articulated objects, we randomly sample joint angles within their valid range (from minimum to maximum limits) in Isaac Sim to induce diverse motions; for assets with pre-defined animations, we vary the playback speed to broaden motion distributions and increase dynamic diversity. We additionally introduce a dedicated hand–object interaction subset. In Blender, we import object meshes along with their motion trajectories, and synthesize interaction motions using the MANO hand model~\cite{Mano} with high-fidelity hand textures~\cite{qian2020html}. We then deploy a multi-view camera setup for rendering and data capture, producing point tracking annotations for hand–object interactions.

To further improve diversity and generalization, we apply data augmentation to the constructed scenes by randomizing camera, lighting, and appearance factors. Specifically, we perturb the camera’s initial pose and motion trajectories with Gaussian noise and randomize the field of view (FOV) to induce both close-up and long-range viewpoints; we vary light source positions and intensities to cover diverse illumination conditions; and we replace textures/materials of selected objects to enrich appearance variations, for example, in the hand–object interaction subset, we sample high-quality hand textures from HTML~\cite{qian2020html} to replace the default MANO textures, enhancing realism and visual diversity. We show representative samples in Fig.~\ref{fig:da}.

\subsection{Data Annotation}
A key feature of our dataset is the cross-platform data generation and annotation design. We support building sequences from both Blender and Isaac Sim, enabling flexible content creation across different application domains. In particular, Isaac Sim is widely adopted in embodied AI and robotics, allowing us to naturally incorporate interactive and physics-based environments. Importantly, all parts are exported in a unified format with consistent definitions, making the dataset scalable and easy to extend to new simulators. Our dataset consists of the following components:


\noindent \textbf{RGB video.} For each camera view, we render a temporally consistent RGB frame sequence. These frames are obtained via standard rendering from the simulator engine with fixed camera parameters.

\noindent \textbf{Depth map.} We provide depth maps aligned with RGB, offering geometric cues for both 2D and 3D reasoning. Depth is obtained from the renderer's depth output (e.g., Z-buffer or distance-to-camera pass) under the same camera projection.

\noindent \textbf{Instance mask.} Pixel-wise instance-level maps are provided for category-aware analysis. Instance labels are generated via simulator-supported segmentation passes, typically based on object identifiers.

\noindent \textbf{Camera pose.} For each frame and view, we record camera intrinsics and pose. Camera parameters are directly exported from the simulator, including intrinsic calibration and extrinsic transformations (camera to world) in the world frame.

\noindent \textbf{Point trajectory.} We annotate dense point trajectories defined on 3D scene surfaces. Point trajectories are obtained by tracking surface points (e.g., mesh vertices or sampled points) in 3D over time and projecting them onto each view to achieve the 2D point trajectories.

\noindent \textbf{Point visible.} To support occlusion-aware learning and evaluation, we additionally provide point visibility annotations for each frame and view. Visibility is derived from both geometric and semantic consistency, indicating whether a tracked point is visible or occluded in the corresponding observation. In practice, we combine frustum checking, depth consistency against rendered depth maps, and agreement with instance segmentation.

\begin{table*}[ht]
\centering
\caption{SynthVerse Benchmark. For 2D trackers CoTracker series, we lift predicted 2D points to 3D using (GT depth) and (GT camera pose)}
\footnotesize
\setlength{\tabcolsep}{1.6pt}
\renewcommand{\arraystretch}{1.16}

\resizebox{\textwidth}{!}{%
\begin{tabular}{l|ccccc|ccccc|ccccc|ccccc}
\toprule
\multirow{2}{*}{\textbf{Methods}} &
\multicolumn{5}{c|}{\textbf{SynthVerse-Nav}} &
\multicolumn{5}{c|}{\textbf{SynthVerse-Human}} &
\multicolumn{5}{c|}{\textbf{SynthVerse-Animal}} &
\multicolumn{5}{c}{\textbf{SynthVerse-Objects}} \\
& $\mathrm{AJ}_{3D}\uparrow$ & $\mathrm{APD}_{3D}\uparrow$ & $\mathrm{AJ}_{2D}\uparrow$ & $\mathrm{APD}_{2D}\uparrow$ & $\mathrm{OA}\uparrow$ &
  $\mathrm{AJ}_{3D}\uparrow$ & $\mathrm{APD}_{3D}\uparrow$ & $\mathrm{AJ}_{2D}\uparrow$ & $\mathrm{APD}_{2D}\uparrow$ & $\mathrm{OA}\uparrow$ &
  $\mathrm{AJ}_{3D}\uparrow$ & $\mathrm{APD}_{3D}\uparrow$ & $\mathrm{AJ}_{2D}\uparrow$ & $\mathrm{APD}_{2D}\uparrow$ & $\mathrm{OA}\uparrow$ &
  $\mathrm{AJ}_{3D}\uparrow$ & $\mathrm{APD}_{3D}\uparrow$ & $\mathrm{AJ}_{2D}\uparrow$ & $\mathrm{APD}_{2D}\uparrow$ & $\mathrm{OA}\uparrow$ \\
\midrule
CoTracker~\cite{karaev2024cotracker} &
19.6 & 24.1 & \textbf{38.0} & 43.7 & 62.3 &
25.2 & 24.8 & 51.0 & 68.0 & \textbf{84.4} &
12.6 & 18.3 & 49.7 & 66.2 & 76.3 &
28.9 & 38.3 & \textbf{46.1} & 60.2 & \textbf{82.4} \\

CoTracker3~\cite{cotracker3} &
18.7 & 23.4 & 37.0 & 45.4 & 64.8 &
27.1 & 39.2 & \textbf{53.1} & \textbf{74.8} & 83.3 &
14.0 & 21.0 & 53.6 & \textbf{73.5} & 76.8 &
26.7 & 39.4 & 39.2 & 57.8 & 63.4 \\

SpatialTrackerV2-offline~\cite{xiao2025spatialtrackerv2} &
19.4 & 21.1 & 37.5 & 45.5 & 74.3 &
17.7 & 25.5 & 30.6 & 43.7 & 62.1 &
21.2 & 30.1 & 42.5 & 55.5 & 69.3 &
\textbf{30.4} & \textbf{43.4} & 25.7 & 34.2 & 55.0 \\

SpatialTrackerV2-online~\cite{xiao2025spatialtrackerv2} &
\textbf{20.5} & 23.5 & \textbf{38.0} & 46.6 & 78.8 &
22.8 & 30.1 & 40.2 & 53.2 & 81.7 &
23.3 & 31.8 & 48.7 & 59.4 & \textbf{86.4} &
30.1 & 37.0 & 30.9 & 40.3 & 59.1 \\

TAPIP3D-camera~\cite{zhang2025tapip3d} &
12.3 & 22.5 & 31.2 & 52.4 & 82.8 &
\textbf{28.4} & 42.4 & 45.3 & 66.7 & 81.0 &
43.0 & \textbf{56.8} & \textbf{54.4} & 72.9 & 82.0 &
21.4 & 38.8 & 33.2 & \textbf{61.8} & 65.4 \\

TAPIP3D-world~\cite{zhang2025tapip3d} &
13.6 & \textbf{25.1} & 33.0 & \textbf{54.9} & \textbf{82.9} &
28.1 & \textbf{42.5} & 44.7 & 66.9 & 81.2 &
\textbf{43.2} & 56.6 & 54.3 & 72.5 & 82.2 &
21.8 & 38.7 & 33.3 & 61.4 & 66.3 \\

\bottomrule
\end{tabular}%
}


\resizebox{\textwidth}{!}{%
\begin{tabular}{l|ccccc|ccccc|ccccc|ccccc}
\toprule
\multirow{2}{*}{\textbf{Methods}} &
\multicolumn{5}{c|}{\textbf{SynthVerse-Embodied}} &
\multicolumn{5}{c|}{\textbf{SynthVerse-Film}} &
\multicolumn{5}{c|}{\textbf{SynthVerse-Interaction}} &
\multicolumn{5}{c}{\textbf{SynthVerse-mAverage}} \\
& $\mathrm{AJ}_{3D}\uparrow$ & $\mathrm{APD}_{3D}\uparrow$ & $\mathrm{AJ}_{2D}\uparrow$ & $\mathrm{APD}_{2D}\uparrow$ & $\mathrm{OA}\uparrow$ &
  $\mathrm{AJ}_{3D}\uparrow$ & $\mathrm{APD}_{3D}\uparrow$ & $\mathrm{AJ}_{2D}\uparrow$ & $\mathrm{APD}_{2D}\uparrow$ & $\mathrm{OA}\uparrow$ &
  $\mathrm{AJ}_{3D}\uparrow$ & $\mathrm{APD}_{3D}\uparrow$ & $\mathrm{AJ}_{2D}\uparrow$ & $\mathrm{APD}_{2D}\uparrow$ & $\mathrm{OA}\uparrow$ &
  $\mathrm{AJ}_{3D}\uparrow$ & $\mathrm{APD}_{3D}\uparrow$ & $\mathrm{AJ}_{2D}\uparrow$ & $\mathrm{APD}_{2D}\uparrow$ & $\mathrm{OA}\uparrow$ \\
\midrule
CoTracker~\cite{karaev2024cotracker} &
22.9 & 27.3 & 35.8 & 41.3 & 56.9 &
4.7 & 7.3 & 51.4 & 65.3 & 83.2 &
22.6 & 43.6 & 45.0 & 81.8 & 58.5 &
19.5 & 26.2 & 45.3 & 60.9 & 72.0 \\

CoTracker3~\cite{cotracker3} &
\textbf{37.1} & \textbf{44.7} & \textbf{50.6} & \textbf{58.3} & 58.7 &
5.0 & 7.8 & 50.1 & 63.2 & \textbf{83.5} &
20.4 & 38.6 & 46.5 & 82.5 & 59.4 &
21.3 & 30.6 & 47.2 & 65.1 & 70.0 \\

SpatialTrackerV2-offline~\cite{xiao2025spatialtrackerv2} &
25.5 & 30.2 & 42.9 & 47.9 & 78.0 &
9.8 & 15.0 & 48.3 & 59.2 & 78.7 &
13.3 & 24.6 & 31.3 & 49.5 & 55.6 &
19.6 & 27.1 & 37.0 & 47.9 & 67.6 \\

SpatialTrackerV2-online~\cite{xiao2025spatialtrackerv2} &
25.2 & 31.8 & 43.5 & 51.4 & \textbf{82.3} &
10.2 & 15.3 & 56.0 & 66.6 & 83.1 &
23.5 & 31.7 & 42.6 & 54.9 & 59.1 &
22.2 & 28.7 & 42.8 & 53.2 & 75.8 \\

TAPIP3D-camera~\cite{zhang2025tapip3d} &
24.8 & 32.2 & 36.9 & 52.2 & 79.0 &
\textbf{41.9} & \textbf{53.1} & 56.9 & 71.7 & 81.5 &
\textbf{61.5} & \textbf{77.8} & \textbf{71.7} & \textbf{90.4} & \textbf{84.2} &
\textbf{33.3} & 46.2 & 47.1 & 66.9 & 79.4 \\

TAPIP3D-world~\cite{zhang2025tapip3d} &
24.5 & 31.5 & 37.4 & 52.2 & 79.4 &
\textbf{41.9} & 52.7 & \textbf{57.6} & \textbf{72.3} & 82.3 &
60.1 & 77.5 & 71.4 & 90.2 & 84.0 &
\textbf{33.3} & \textbf{46.4} & \textbf{47.4} & \textbf{67.2} & \textbf{79.8} \\

\bottomrule
\end{tabular}%
}

\label{tab:synthverse_all}
\end{table*}

\begin{table*}
\centering
\caption{Validation on public synthetic data benchmark. ($^*$) stands for the results of the finetune model.}
\footnotesize
\setlength{\tabcolsep}{1.3pt}
\renewcommand{\arraystretch}{1.18}

\resizebox{\textwidth}{!}{%
\begin{tabular}{l|ccccc|ccccc|ccccc|ccccc}
\toprule
\multirow{2}{*}{\textbf{Methods}} &
\multicolumn{5}{c|}{\textbf{SynthVerse-mAverage}} &
\multicolumn{5}{c|}{\textbf{Dynamic Replica}~\cite{karaev2023dynamicstereo}} &
\multicolumn{5}{c|}{\textbf{LSFOdyssey}~\cite{wang2025scenetracker}} &
\multicolumn{5}{c}{\textbf{Average}} \\
& $\mathrm{AJ}_{3D}\uparrow$ & $\mathrm{APD}_{3D}\uparrow$ & $\mathrm{AJ}_{2D}\uparrow$ & $\mathrm{APD}_{2D}\uparrow$ & $\mathrm{OA}\uparrow$ &
  $\mathrm{AJ}_{3D}\uparrow$ & $\mathrm{APD}_{3D}\uparrow$ & $\mathrm{AJ}_{2D}\uparrow$ & $\mathrm{APD}_{2D}\uparrow$ & $\mathrm{OA}\uparrow$ &
  $\mathrm{AJ}_{3D}\uparrow$ & $\mathrm{APD}_{3D}\uparrow$ & $\mathrm{AJ}_{2D}\uparrow$ & $\mathrm{APD}_{2D}\uparrow$ & $\mathrm{OA}\uparrow$ &
  $\mathrm{AJ}_{3D}\uparrow$ & $\mathrm{APD}_{3D}\uparrow$ & $\mathrm{AJ}_{2D}\uparrow$ & $\mathrm{APD}_{2D}\uparrow$ & $\mathrm{OA}\uparrow$ \\
\midrule

TAPIP3D-camera~\cite{zhang2025tapip3d} &
33.3 & 46.2 & 47.1 & 66.9 & 79.4 &
53.7 & 70.8 & 64.6 & 84.7 & 84.7 &
68.3 & \textbf{83.2} & 76.0 & \textbf{91.2} & 86.2 &
51.8 & 66.7 & 62.6 & 80.9 & 83.4 \\

\rowcolor{gray!20} TAPIP3D-camera$^{*}$ &
\textbf{41.8} & \textbf{51.7} & \textbf{52.3} & \textbf{68.6} & \textbf{82.9} &
\textbf{56.6} & \textbf{74.3} & \textbf{66.4} & \textbf{85.7} & \textbf{86.0} &
\textbf{70.8} & 82.8 & \textbf{79.0} & 91.0 & \textbf{91.1} &
\textbf{56.4} & \textbf{69.6} & \textbf{65.9} & \textbf{81.8} & \textbf{86.7} \\

\midrule
TAPIP3D-world~\cite{zhang2025tapip3d} &
33.3 & 46.4 & 47.4 & 67.2 & 79.8 &
55.5 & 72.8 & 66.2 & 85.7 & 85.3 &
72.2 & \textbf{85.8} & 78.5 & 92.8 & 86.9 &
53.7 & 68.3 & 64.0 & 81.9 & 84.0 \\

\rowcolor{gray!20} TAPIP3D-world$^{*}$ &
\textbf{41.6} & \textbf{52.1} & \textbf{52.7} & \textbf{69.1} & \textbf{82.9} &
\textbf{57.2} & \textbf{74.5} & \textbf{67.3} & \textbf{86.4} & \textbf{86.1} &
\textbf{73.9} & 85.5 & \textbf{81.5} & \textbf{92.9} & \textbf{91.2} &
\textbf{57.6} & \textbf{70.7} & \textbf{67.2} & \textbf{82.8} & \textbf{86.7} \\

\bottomrule
\end{tabular}%
}
\label{tab:public_synth}
\end{table*}

\section{Experiment}

\subsection{Experiment Setting}

\noindent \textbf{Implement details.} We perform large-scale batch rendering of our data on a cluster equipped with 200 NVIDIA RTX 4090 GPUs (48GB each), and fine-tune TAPIP3D using 8 NVIDIA H200 GPUs (141GB each). We primarily use Blender 4.2 and Isaac Sim 4.5 for scene construction and rendering. A small portion of animated-film projects requires Blender 3.x for rendering due to compatibility with legacy production files. For fine-tuning TAPIP3D, we adopt a OneCycle learning rate schedule with an initial learning rate of 1e-4 and a total of 50K training steps.

\noindent \textbf{Evaluation Metric.} Following TAPIP3D~\cite{zhang2025tapip3d}, we adopt  $\mathrm{AJ}_{3D}$, $\mathrm{APD}_{3D}$, $\mathrm{AJ}_{2D}$, $\mathrm{APD}_{2D}$, $\mathrm{OA}$ as the main evaluation metric. $\mathrm{AJ}_{3D}$ assesses tracking rigor by calculating the spatial and visibility overlap between predicted points and ground truth in 3D coordinates. $\mathrm{AJ}_{3D}$ quantifies the proportion of points whose predicted 3D positions fall within a depth-adaptive error tolerance relative to the actual coordinates. In addition, $\mathrm{OA}$ specifically evaluates the accuracy of the model's binary classification regarding whether a tracked point is occluded or out of frame. After mapping results back to 2D images, $\mathrm{AJ}_{2D}$ measures the spatio-temporal alignment of pixels across video sequences using the Intersection-over-Union metric. Moreover, $\mathrm{APD}_{2D}$ measures whether the average Euclidean distance between the predicted trajectories and ground truth on the pixel plane remains within minimal error bounds.

\subsection{SynthVerse Benchmark}

Our SynthVerse benchmark consists of eight subsets spanning diverse domains and scene configurations (i.e., Nav, Human, Animal, Objects, Embodied, Film, Interaction, and the overall mAverage). For each subset, we report five metrics, including $\mathrm{AJ}{2D}$, $\mathrm{APD}{2D}$, $\mathrm{AJ}{3D}$, $\mathrm{APD}{3D}$, and occlusion accuracy ($\mathrm{OA}$), to provide a comprehensive evaluation of tracking quality. Specifically, we evaluate representative state-of-the-art 2D and 3D point tracking methods on our proposed benchmark, including CoTracker~\cite{karaev2024cotracker}, CoTracker3~\cite{cotracker3}, SpatialTrackerV2~\cite{xiao2025spatialtrackerv2}, and TAPIP3D~\cite{zhang2025tapip3d}. The quantitative results are summarized in Table~\ref{tab:synthverse_all}.

Across different subsets, Nav and Film are consistently more challenging for all methods. This is likely due to rapid viewpoint changes, large camera motion, and fast scene transitions, which significantly increase correspondence ambiguity and occlusion frequency. Moreover, such dynamics are still under-represented in existing training corpora, making current trackers less robust under these settings. In contrast, Interaction and Animal are relatively easier for current methods, as they often exhibit more structured motion patterns and more common visual appearances, leading to more stable tracking performance. These diverse difficulty levels across subsets further reflect the broad coverage and diversity of SynthVerse.

\begin{figure}

\centering
\includegraphics[width=0.65\textwidth]{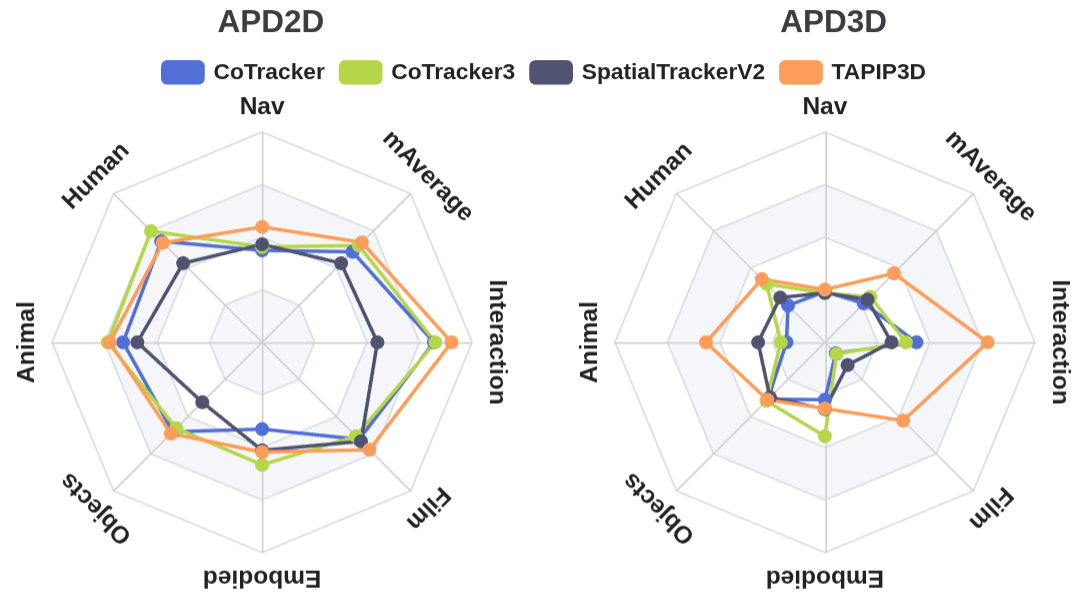} 
\caption{SynthVerse Benchmark Radar. For simplicity, we denote SpatialTrackerV2-online as SpatialTrackerV2 and refer to TAPIP3D-world as TAPIP3D.}
\label{fig:lindar}
\end{figure}

Furthermore, we visualize the domain-wise $\mathrm{APD}_{2D}$ and $\mathrm{APD}_{3D}$ using radar plots in Fig.~\ref{fig:lindar}. A clear trend emerges that 3D point tracking remains substantially more challenging than its 2D counterpart: while existing methods achieve relatively stable performance in 2D across most subsets, their 3D accuracy exhibits a much larger degradation and variance. More importantly, the radar plots reveal pronounced domain-dependent behaviors, where the relative ranking and performance gaps change noticeably across Nav, Film, Embodied, and Interaction scenarios. This observation highlights that current trackers are still sensitive to distribution shifts in scene type, motion pattern, and occlusion structure, and no single method performs uniformly well across all domains. These results further validate the importance of our diverse benchmark.

\subsection{Model Fine-tuning and Efficacy Validation}
We adopt TAPIP3D as the primary model for fine-tuning and validation, as it is a competitive and representative open-source approach for 2D/3D point tracking. Importantly, TAPIP3D provides publicly available training and evaluation code, enabling reproducible experiments and fair comparisons under a unified setup. In addition, we validate the effectiveness and generalization of SynthVerse from three perspectives: (i) evaluation on our proposed SynthVerse benchmark, (ii) external validation on public synthetic datasets to assess cross-dataset generalization, and (iii) evaluation on public realistic benchmarks to examine transferability to real-world scenarios.

\noindent \textbf{On our SynthVerse benchmark} We first analyze the effect of fine-tuning on our SynthVerse benchmark. As shown in Table~\ref{tab:public_synth}, fine-tuning on SynthVerse leads to a substantial improvement on SynthVerse-mAverage for both TAPIP3D variants, with $\mathrm{AJ}_{3D}$ increasing from $33.3$ to $41.8$ (camera) and from $33.3$ to $41.6$ (world), accompanied by consistent gains in $\mathrm{APD}_{3D}$ and $\mathrm{OA}$. These results indicate that SynthVerse provides strong and diverse supervision signals, effectively enhancing 3D tracking robustness under domain shifts within our benchmark.

\begin{figure}[ht]

\centering
\includegraphics[width=0.7275\textwidth]{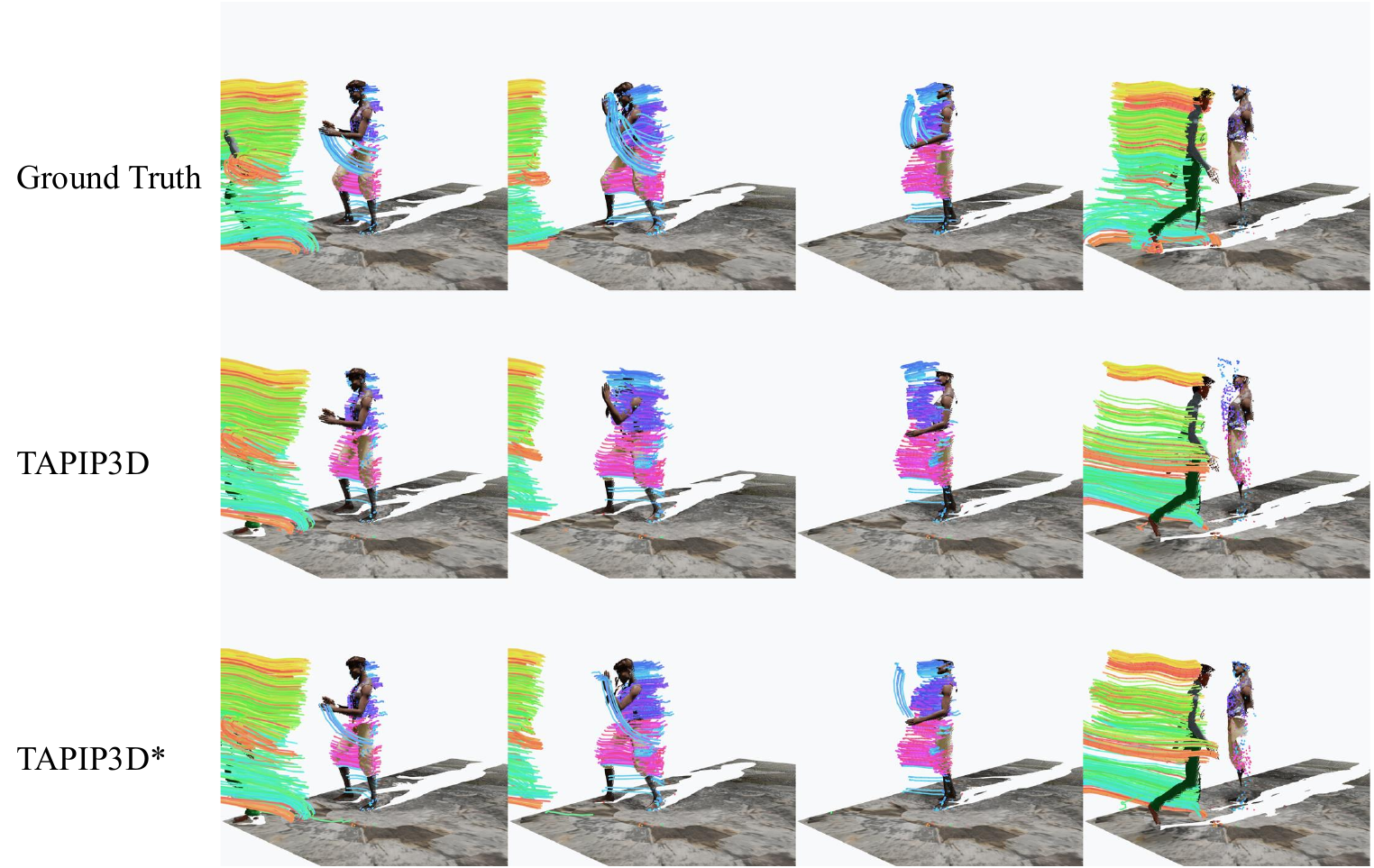} 
\caption{Qualitative Comparison on SynthVerse Benchmark.  For simplicity, we denote TAPIP3D-world as TAPIP3D and refer to TAPIP3D-world$^{*}$ as TAPIP3D$^{*}$.  ($^*$) stands for the results of the finetune model.}
\label{fig:qualitative_comparison}
\end{figure}

Fig.~\ref{fig:qualitative_comparison} illustrates a qualitative comparison on our SynthVerse benchmark. Compared to the ground truth, the baseline TAPIP3D exhibits two prominent failure modes. First, it fails to maintain stable correspondences on fine-grained and fast-moving regions, resulting in missing or prematurely terminated trajectories, which is particularly evident around the hand where rapid articulation and partial occlusions are frequent. Second, even when trajectories are preserved, TAPIP3D often suffers from noticeable localization errors: the predicted tracks gradually drift away from the correct motion pattern and become misaligned across frames under large motions and occlusions. After fine-tuning on SynthVerse (TAPIP3D$^{*}$), both issues are mitigated to a large extent, with better preservation of hand trajectories and reduced drift, indicating improved robustness to challenging dynamics and domain shifts.

\noindent  \textbf{On public synthetic benchmark.} We further examine the transferability of SynthVerse fine-tuning on public synthetic benchmarks, Dynamic Replica and LSFOdyssey. As shown in Table~\ref{tab:public_synth}, fine-tuning yields consistent improvements on both datasets for both TAPIP3D variants. Compared to the gains observed on our SynthVerse benchmark, the improvements on these public synthetic datasets are relatively moderate, which is expected since TAPIP3D is originally trained on data distributions closely related to Dynamic Replica and LSFOdyssey. Nevertheless, the average performance over all three synthetic benchmarks also increases accordingly, confirming that SynthVerse provides complementary supervision that benefits generalization beyond the proposed benchmark.

\noindent  \textbf{On public realistic benchmark.} We further validate the sim-to-real transferability on three public realistic benchmarks, including ADT, DriveTrack, and PStudio, as summarized in Table~\ref{tab:public_real}. Fine-tuning on SynthVerse consistently improves performance on all three datasets, demonstrating that the learned robustness can generalize beyond synthetic environments to real-world observations. We note that these realistic benchmarks focus on relatively narrow scene types and content distributions compared to SynthVerse; nevertheless, the observed gains indicate that SynthVerse provides complementary supervision that helps mitigate the domain gap and enhances tracking robustness under real-world motion and occlusion patterns.

\begin{table}
\centering
\caption{Validation on public realistic data benchmark. ($^*$) stands for the results of the finetune model.}
\footnotesize
\setlength{\tabcolsep}{1.5pt}
\renewcommand{\arraystretch}{1.18}

\resizebox{0.7775\textwidth}{!}{%
\begin{tabular}{l|ccc|ccc|ccc}
\toprule
\multirow{2}{*}{\textbf{Methods}} &
\multicolumn{3}{c|}{\textbf{ADT}~\cite{pan2023aria}} &
\multicolumn{3}{c|}{\textbf{DriveTrack}~\cite{balasingam2024drivetrack}} &
\multicolumn{3}{c}{\textbf{PStudio}~\cite{joo2015panoptic}} \\
& $\mathrm{AJ}_{3D}\uparrow$ & $\mathrm{APD}_{3D}\uparrow$ & $\mathrm{OA}\uparrow$ &
  $\mathrm{AJ}_{3D}\uparrow$ & $\mathrm{APD}_{3D}\uparrow$ & $\mathrm{OA}\uparrow$ &
  $\mathrm{AJ}_{3D}\uparrow$ & $\mathrm{APD}_{3D}\uparrow$ & $\mathrm{OA}\uparrow$ \\
\midrule

TAPIP3D-camera~\cite{zhang2025tapip3d} &
21.6 & 31.0 & 90.4 &
14.6 & 21.3 & 82.2 &
18.1 & 27.7 & 85.5 \\

\rowcolor{gray!20} TAPIP3D-camera$^{*}$ &
\textbf{22.9} & \textbf{31.8} & \textbf{93.2} &
\textbf{15.1} & \textbf{21.6} & \textbf{85.4} &
\textbf{18.6} & \textbf{28.2} & \textbf{87.2} \\

\midrule
TAPIP3D-world~\cite{zhang2025tapip3d} &
23.5 & 32.8 & 91.2 &
14.9 & 21.8 & 82.6 &
18.1 & 27.7 & 85.5 \\

\rowcolor{gray!20} TAPIP3D-world$^{*}$ &
\textbf{24.3} & \textbf{33.3} & \textbf{93.5} &
\textbf{15.4} & \textbf{22.2} & \textbf{85.3} &
\textbf{18.6} & \textbf{28.2} & \textbf{87.2} \\

\bottomrule
\end{tabular}%
}

\label{tab:public_real}
\end{table}

Overall, fine-tuning TAPIP3D on SynthVerse consistently improves performance not only on our proposed benchmark but also across five widely used public datasets, demonstrating the effectiveness of SynthVerse as a scalable training source. Importantly, these public benchmarks span different domains and content distributions, ranging from synthetic to realistic scenarios. The consistent gains observed across such heterogeneous evaluations suggest that the diverse coverage of SynthVerse provides complementary supervision that enhances robustness under domain shifts, benefiting general point tracking across multiple domains.

\section{Conclusion}
In this work, we present SynthVerse, a large-scale and diverse synthetic dataset and benchmark for general 2D/3D point tracking, built upon a unified cross-platform generation pipeline integrating Blender and Isaac Sim. SynthVerse provides broad domain and object coverage with reliable trajectory and visibility annotations, enabling systematic training and evaluation under diverse motions, occlusions, and distribution shifts. Extensive experiments demonstrate that SynthVerse serves as an effective resource for improving robustness and generalization, while also revealing limitations of existing trackers across different domains. In future work, we will benchmark additional state-of-the-art models on SynthVerse to further characterize current progress.

\bibliographystyle{unsrt}  
\bibliography{references}

\end{document}